\documentclass{article}
\usepackage[final]{corl_2017}

\usepackage{xspace}
\usepackage{hyperref}
\usepackage{pdfsync}

\newcommand{\control}{\ensuremath{F}\xspace}
\newcommand{\BLIND}{\textrm{Path Tracker}\xspace}
\newcommand{\PURE}{\textrm{Non-Intention Net}\xspace}
\newcommand{\DWA}{\textrm{Dynamic Window}\xspace}
\newcommand{\IN}{\textrm{DLM-Net}\xspace}
\newcommand{\IMN}{\textrm{LPE-Net}\xspace}
\renewcommand{\path}{\ensuremath{\sigma}\xspace}
\newcommand{\cmd}[1]{{\small \textsc{#1}}}
\newcommand{\pos}{\ensuremath{x}\xspace}

\usepackage{subfig}
\usepackage{amsmath}
\usepackage[font=small]{caption}
\title{Intention-Net:  Integrating Planning and Deep Learning for Goal-Directed Autonomous Navigation}

\author{
  Wei Gao\qquad\qquad David Hsu \qquad\qquad Wee Sun Lee\\
  School of Computing\\
  National University of Singapore\\
  \texttt{\{gaowei90, dyhsu, leews\}@comp.nus.edu.sg} 
  \And
  Shengmei Shen\qquad\quad Karthikk Subramanian\\
  Panasonic R\&D Center Singapore\\
  \texttt{\{shengmei.shen, karthikk.subramanian\}@sg.panasonic.com} \\
}

\usepackage{amsmath,amssymb}

\usepackage{wrapfig}
\usepackage{graphicx}
\graphicspath{{figures/}}
\DeclareGraphicsExtensions{.pdf,.jpg,.png,.eps}

\newcommand{\secref}[1]{Section~\ref{#1}}
\renewcommand{\eqref}[1]{(\ref{#1})}
\newcommand{\figref}[1]{Figure~\ref{#1}}
\newcommand{\subfig}[1]{{\textrm{#1}}}
\newcommand{\tabref}[1]{Table~\ref{#1}}

\newcommand{\ie}{\textit{i.e.}}
\newcommand{\eg}{\textit{e.g.}}
\newcommand{\etc}{\textit{etc.}}
\newcommand{\etal}{\textit{et~al.}}

\newlength{\citeskipup}
\newlength{\citeskipdown}

\usepackage{color}
\definecolor{fullred}{rgb}{0.95,0.0,0.1} 
\newcounter{cmt}

\begin{document}
\maketitle


\begin{abstract}
  How can a delivery robot navigate reliably to a destination in a new
  office building, with minimal prior information?
  To tackle this challenge, this
  paper introduces a two-level hierarchical approach, which integrates
  model-free deep learning and model-based path planning.  At the low level,
  a neural-network motion controller, called the  \emph{intention-net}, 
  is trained end-to-end to provide robust local navigation.
  The intention-net maps images
  from a single monocular camera and  ``intentions'' directly to robot
  controls.
  At the high level, a path planner uses a crude map, \eg, a 2-D floor plan,
  to compute a  path from the robot's current location to the goal. The
  planned path  provides intentions to the intention-net.
  Preliminary experiments suggest that
  the learned motion controller is robust against perceptual uncertainty and
  by integrating with a path planner, it generalizes
  effectively to  new environments and goals.
\end{abstract}



\section{Introduction}
Goal-directed, collision-free global navigation is an essential capability of
autonomous robots. However, robots still do not quite match humans in such
navigation tasks, despite decades of research and a vast
literature~\cite{bonin2008visual,fallah2013indoor}.
For example, upon arriving at
a new shopping complex, a human can follow an abstract 2-D floor plan,
devoid of most geometric and visual details of the 3-D world, and reach
any location in the building, unimpeded by railings, glass walls, \ldots that
often cause failures in robot navigation systems.  Similarly, a human can
drive to any destination in a new city by following a roadmap or GPS based
navigation directions.  There are two key elements to humans' performance: the
ability to plan a path using a simplified abstract model of the 3-D world and
more importantly, the ability to execute the planned path robustly using local
visual information.  To achieve comparable performance in robot navigation, we
propose to integrate model-free deep learning for local collision avoidance
and model-based path planning for goal-directed global navigation.

The idea of combining global path planning and local motion control is well
studied in collision-free robot navigation~\cite{bruce2002real,
  desouza2002vision}. There are many path planning methods~\cite{Lav06}.
There are many motion control methods as well, \eg,
potential-field~\cite{koren1991potential}, visual
servoing~\cite{cowan2002visual}, \ldots, but they are usually manually
designed.  In
recent years, deep learning has achieved extraordinary success in many
domains,  from image classification, speech recognition to game
playing~\cite{he2016deep, hinton2012deep, silver2016mastering}.  Deep learning
has also found application in navigation tasks. Earlier work, however, usually
trains
the agent to move along a fixed path~\cite{bojarski2016end}, with some notable
exceptions~\cite{chen2015deepdriving}.

\begin{figure}[t]
	\centering
    \subfloat[(a)]{\label{workflow:a}\includegraphics[width=0.155\textwidth]{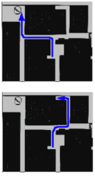}}\,
    \subfloat[(b)]{\label{workflow:b}\includegraphics[width=0.40\textwidth]{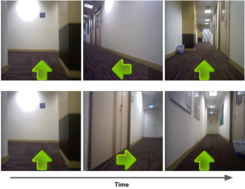}}\,
    \subfloat[(c)]{\label{workflow:c}\includegraphics[width=0.40\textwidth]{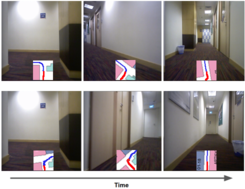}}\,
    \caption{A two-level navigation hierarchy,
      with a path planner at the top and the \emph{intention-net},
      a learned local motion controller, at the bottom. (\subfig a) Two paths
      planned for different goals in the same environment.
      (\subfig b) The intention as a discretized local move.
      (\subfig c) The intention as a local path and environment.}
	\label{fig:workflow}
\end{figure}

Our proposed approach consists of a path planner at the high level and a
motion controller, trained via deep learning, at the low level
(\figref{fig:workflow}).  The path
planner computes a collision-free global path according to a crude input map,
\eg, a 2-D floor plan.  The motion controller tracks this path, using images
from a single monocular camera onboard the robot.  It handles all local
environment dynamics, including obstruction from static and dynamic obstacles
not in the input map---such as furniture and people---and social conventions
for navigation.  The local motion controller
is trained end-to-end via imitation learning.
After training, the robot  navigates
to arbitrary goals in previously \emph{unseen}
environments.  The proposed two-level architecture is more robust against
perceptual uncertainty than traditional navigation systems, as the motion
controller is learned from data. Further, by integrating with a path planner,
it the learned motion controller generalizes to new
environments and goals  effectively. 

We represent the local motion controller as a deep neural network \control.
Previous work trains \control via imitation learning to follow roads and avoid
obstacles, with impressive performance~\cite{bojarski2016end}.  However, it
implicitly assumes \control to be a
mapping directly from perceptual inputs to controls.
The robot is thus not steerable according to high-level goals.
For example, at an
intersection, the decision to go straight or make a turn depends on not only
the local perceptual input, but also the  navigation goal. Mimicking
the expert in the training dataset does not lead to the right action, as the
goals may be different. Instead, we train a neural-network controller, called
the \emph{intention-net}, conditioned on both perceptual inputs and
\emph{intentions}.  In our two-level architecture, the planned path at the top
provides the intention for the motion controller at the bottom.  The
controller chooses the robot control conditioned on both the intention and the
local perceptual input.
Specifically, the
\emph{intention} may represent a subgoal along the path and the local
environment at the robot's current location.

This work introduces a two-level hierarchical approach for robot navigation,
with a path planner at the top and the intention-net, a motion controller
learned from data, at the bottom (\secref{sec:intention}).
The intention is the key ingredient that binds together the two levels of the
hierarchy. Preliminary results in
simulation and  real-robot experiments suggest that the new approach
enables a mobile
robot to navigate reliably  in a new environment with only a crude map
(\secref{sec:experiments}).


\section{Related Work}\label{sec:related}
Deep learning has been immensely successful in many
domains~\cite{bojarski2016end,chen2015deepdriving,
  krizhevsky2012imagenet,hinton2012deep,levine2016end,sadeghi2016cad,silver2016mastering}.
In robot navigation, one use of deep learning is to learn a flight controller
that maps perceptual inputs directly to control for local collision-free
maneuver of a drone~\cite{sadeghi2016cad}.  It addresses the issue of local
collision avoidance, but not that of goal-directed global navigation. Another
use is to train a system end-to-end for autonomous driving, using monocular
camera images~\cite{bojarski2016end}, but the system drives along a fixed
route and cannot reach arbitrary goals. Some recent work explores model-free
end-to-end learning for goal-directed navigation by incorporating the goal as
part of the perceptual inputs~\cite{pfeiffer2016perception,zhu2016target}. One
may improve the learning efficiency and navigation performance by adding
auxiliary objectives, such as local depth prediction and loop closure
classification~\cite{mirowski2016learning}.  Without a model, these approaches
cannot exploit the sequential decision nature of global navigation effectively
and have difficulty in generalizing to complex new environments.  To tackle
this challenge, our proposed two-level architecture integrates model-free deep
learning for local collision avoidance and model-based global path planning,
using a crude map.

Hierarchies are crucial for coping with computational complexity, both in
learning and in planning. Combining learning and planning in the same
hierarchy is, however, less common.  Kaelbling \etal proposed to learn
composable models of parameterized skills with pre- and post-conditions so
that a high-level planner can compose these skills to complete complex
tasks~\cite{kaelbling2017learning}.
Our hierarchical method shares similar thinking, but specializes to
navigation tasks. It uses intention instead of general pre- and
post-conditions to bind together hierarchical layers and achieves great
efficiency.

Navigation is one of the most important robotic tasks. One classic approach
consists of three steps: build a map of the environment through, \eg,
SLAM~\cite{aulinas2008slam},
plan a path using the map, and finally execute the
path.  High-fidelity geometric maps make it easier for
path planning and execution. However, building such maps is
time-consuming. Further, even small environmental changes may render them
partially invalid.  Alternatively, the optical flow approach does not use maps
at all and relies on the visual perception of the local environment for
navigation~\cite{chao2014survey}.  While this avoids the difficulty of
building and maintaining accurate geometric maps, it is difficult to achieve
effective goal-directed global navigation in complex geometric environments
without maps.  Our approach uses crude maps, \eg, 2-D floor plans, and sits
between the two extremes. Floor plans are widely available for many indoor
environments.  One may also sketch them by hand.
\section{Integrated Planning and Learning with Intention-Net}
\label{sec:intention}

Our proposed hierarchical method performs closed-loop planning and control.
At each time step, the path planner at the high level replans a path from the
robot's current position to the goal, using a crude floor-plan map. The path
is processed to generate the ``intention'' for the intention-net motion
controller. Given the intention and an image from a single monocular camera,
the low-level controller computes the desired speed and steering angle for the
robot to execute. The process then repeats. 
In this section, we first present the intention-net motion controller (\secref{sec:controller}) and the
learning method (\secref{sec:learn}). We then briefly describe the path planner (\secref{sec:plan}). 


\subsection{Intention-Net}\label{sec:controller}
\begin{wrapfigure}[18]{r}{1.6in}
  \vspace{-8pt}
  \centering
  \includegraphics[width=1in]{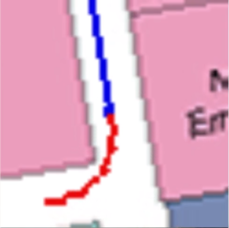}
      \caption{The LPE intention represented as a $224\times 224$ image. It
        captures information on the path that the robot has recently traversed
        (in red), the path ahead (in blue), the robot's current position
        (where the red and the blue paths meet), and
        the local geometric environment according to the map. }
      \label{fig:LPE}
    \end{wrapfigure}
\emph{Intention} binds together global path planning and local motion control.
Intuitively, intention captures what the path planner expects the motion
controller to do at the robot's current position.  We propose two definitions
of intention. The simpler one mimics the instructions usually given for
navigation: go straight, turn left, \etc\,. We call it \emph{discretized local
  move} (DLM). DLM may take four discrete values: \cmd{TurnLeft},
\cmd{TurnRight}, \cmd{GoForward}, and \cmd{Stop}.  Given a path
\path from the robot's current position $\pos$  to the goal,
we compute DLM by estimating the local signed curvature of \path at $\pos$.
If the absolute value of the curvature is smaller
than a chosen threshold, then the DLM is \cmd{GoForward}. Otherwise, the
DLM is either \cmd{TurnLeft} or \cmd{TurnRight} according to the sign of
the curvature. The DLM is \cmd{Stop} if the robot reaches the goal.
DLM is intuitive, but restricts the intention to four discrete values. This is
clearly inadequate, for example, at an intersection with five
or more branches.   
Our second definition, \emph{local path and environment} (LPE), is richer. It
is represented as a $224\times 224$ image,  containing both the path and the
environment within a local window of the current robot position (\figref{fig:LPE}). See
\figref{fig:workflow}c for additional examples.

We represent the intention-net is a deep multivariate regression network
\control.  At time step $t$, it takes as input a camera image $X_t$ and the
intention $I_t$, which is obtained from the path computed by the high-level
planner. It outputs the robot control $\mu_t=(v_t, \theta_t)$, where
$v_t \text{ and }\theta_t$ are the robot speed and steering angle,
respectively. Formally, $\mu_t=(v_t, \theta_t) = \control(X_t, I_t)$.

\begin{figure}[t]
	\centering
    \subfloat[(a)]{\label{architecture:a}\includegraphics[width=0.45\textwidth]{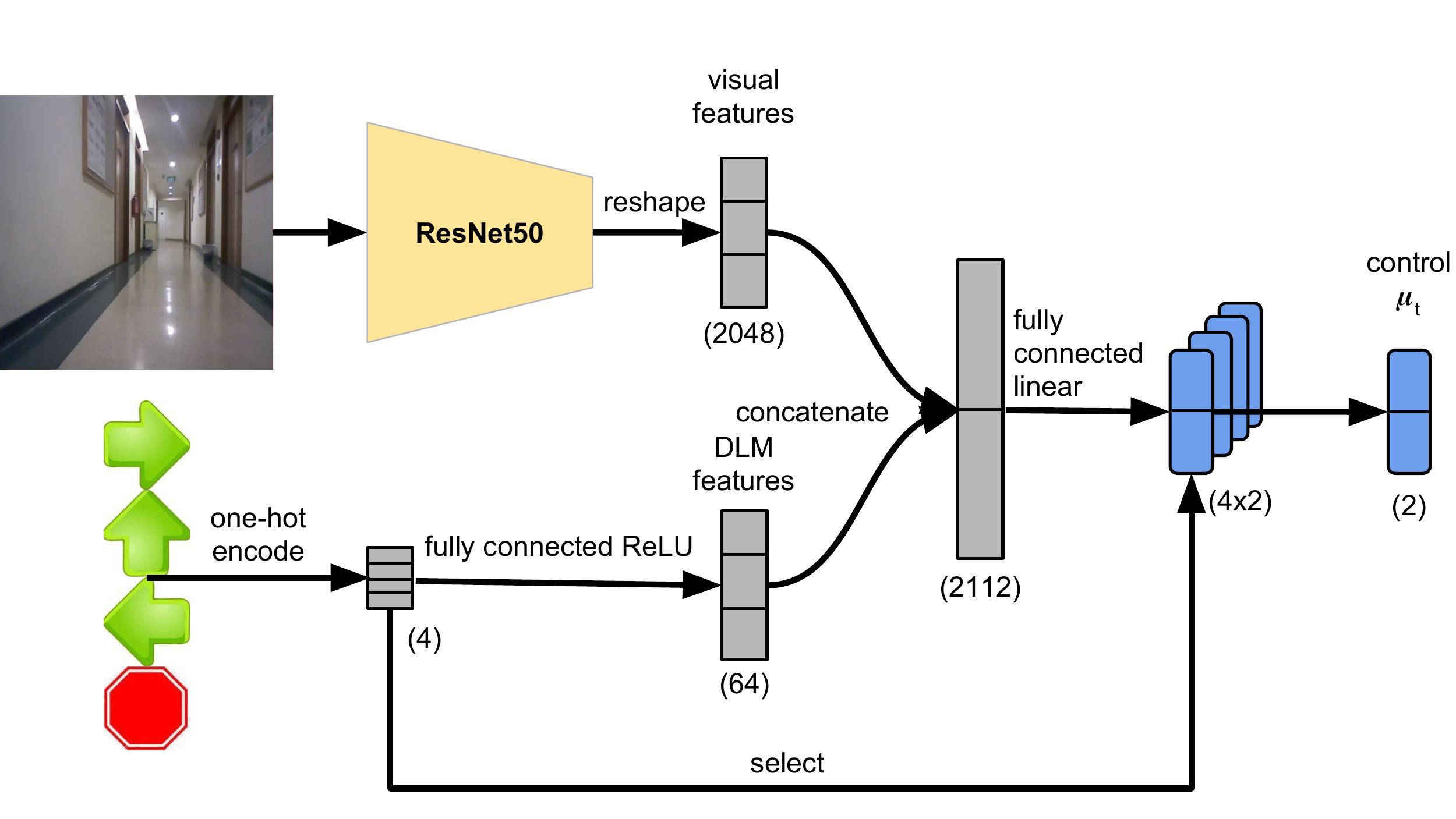}}\quad\quad
    \subfloat[(b)]{\label{architecture:b}\includegraphics[width=0.45\textwidth]{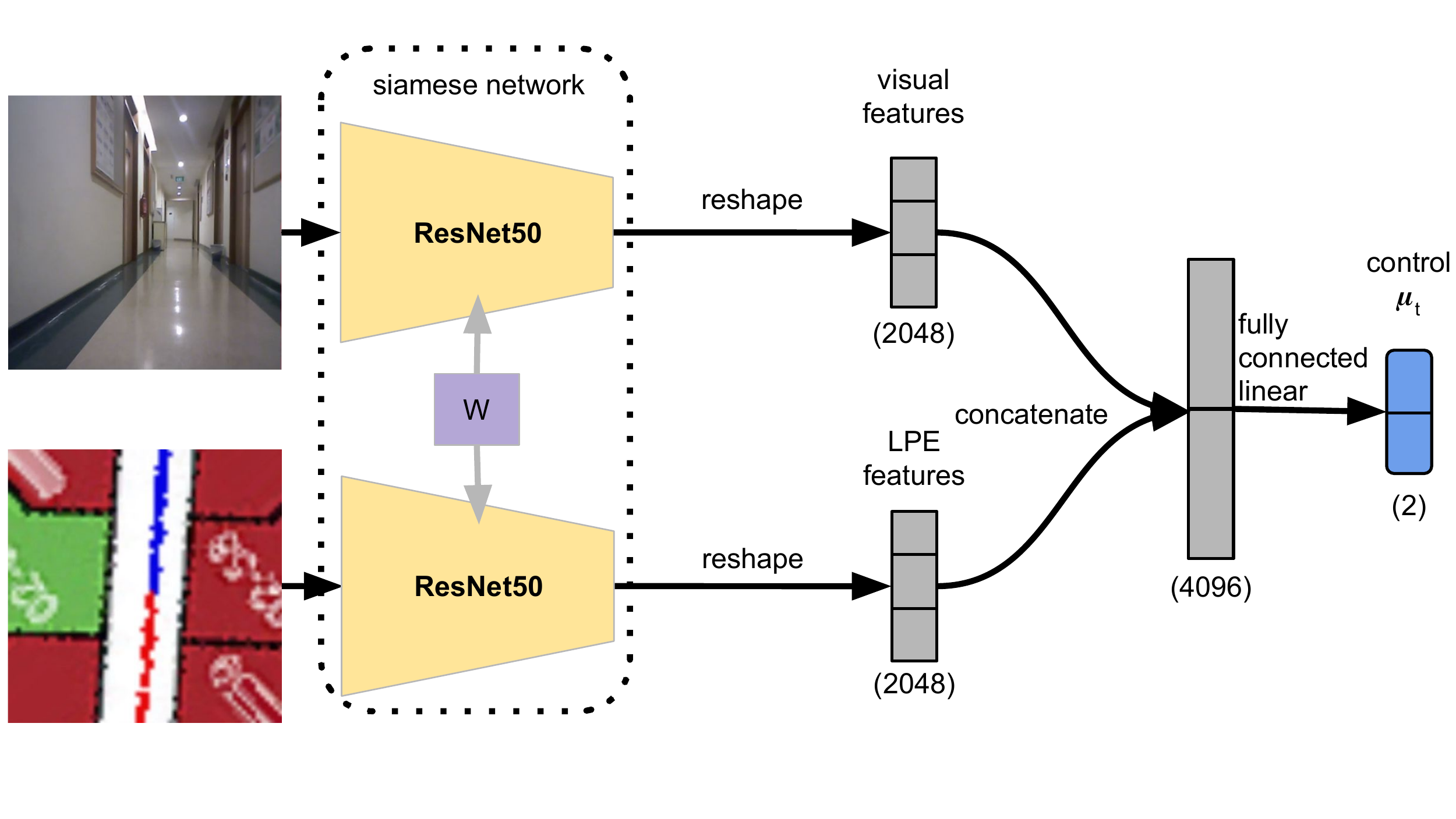}}
    \caption{
      Intention-net architectures. (\subfig a) DLM-net.
      The DLM intention takes four
      discrete values: \textsc{TurnRight}, \mbox{\textsc{GoForward}},
      \textsc{TurnLeft}, and \textsc{Stop}.  (\subfig b) LPE-net.
      The LPE intention is 
      represented as a $224\times 224$ image. Both networks also take a
     $224\times224$ camera image as the input.  
    }
	\label{fig:architecture}
\end{figure}

\figref{fig:architecture} provides a sketch of the network architectures for
the DLM-net and the LPE-net. The DLM-net takes as input a camera image of size
$224\times 224$ and the DLM intention with four discrete values. It extracts
from the image a feature vector of length~2048, using ResNet50 without the
last fully connected layer.  It encodes the intention first as a binary vector
of length~4 and then expands it to a vector of length~64 in order to balance
the influence of the visual and the intention feature vectors.  The visual and
the intention feature vectors are concatenated together as a single vector,
which is then processed by a fully connected linear layer to generate the
controls for all intentions. The DLM-net finally selects the control using the
input intention.


The architecture for the LPE-net is even simpler conceptually.  It takes as
input a camera image and an intention image, both of size $224\times 224$, and
process them through a two-stream siamese network with shared weights.  The
intention image represents the environment according to the map and the robot
path within a local window of the robot's current position. The camera image
captures visual appearance of the 3-D world.  The siamese network tries to
reasoning about the spatial relationships between the two and embeds them in a
space that preserves their spatial relationships. The siamese network outputs
two feature vectors of length~2048. They are concatenated and processed by a
fully connected linear layer to generate the final robot control.


\subsection{Data Collection and Training}\label{sec:learn}

We collect data from demonstrations in the form of tuples
$(X_t, I_t, \mu_t)$ for $t=1,2, \ldots$\,.
For the simulation experiments, we run the path planner to generate the
intention $I_t$ and use the dynamic window approach~\cite{FoxBur97} as the
expert to generate the robot control $\mu_t$.  For the real-robot experiments,
we collect data in a similar way, but use human experts.  We run the path
planner to generate the intention $I_t$ and visualize it. The human expert
controls the robot with a joystick according to the visualized intention.
The collected dataset is split randomly into 4:1 ratio for
training and evaluation.

We learn the desired velocity $v_t$ and steering angle $\theta_t$  jointly and 
reweight their values so that $v_t, \theta_t \in [-1, 1]$, for all $t$.   
We use rmsprop optimizer with initial learning
rate $1e-4$ and L2 weight regularization $1e-4$ to training our network. We
anneal the learning rate over time by $\alpha=\alpha_0/(1+k)$, where $k$ is
the number of iterations and $\alpha_0$ is the initial learning rate. We
observe that training with smaller batch size dramatically increases the
performance of the final regression. For practical use, we use a batch size of
$8$ in our experiments. We also find that using large training samples per
epoch reduces the validation loss significantly in training.
Specifically, we maintain
a dataset pool that can be resampled at any time. We use $200,000$ uniformly
sampled samples from the dataset pool in each epoch. Maintaining the dataset
pool also makes it possible to create  a large number of
samples with a relatively small dataset.




\subsection{Path Planning}\label{sec:plan}
The path planner replans a path from the robot's current
position to the goal at every time step. It represents a crude map as an
occupancy grid and performs adaptive Monte Carlo localization
(AMCL)~\cite{fox2001kld} to localize the robot and determine its current
position with respect to the map.  The planner searches the occupancy grid for
the path using the hybrid $A^*$ algorithm~\cite{DolThr10}, which can
accommodate robot kinematic and dynamic constraints.

The crude map may cause inaccuracies in both localization and path
planning. Our hierarchical approach addresses the issue in two ways.  First,
the planned path is not executed directly. Its primary purpose is to generate
the intention.  The intention-net motion controller combines the intention
with robot's local perceptual input to generate the controls for execution. By
training on suitable datasets, intent-net learns to be robust against some of
these inaccuracies. Second, replanning at each time step further improves
robustness.

\section{Experiments}
\label{sec:experiments}

\begin{figure}[t]
\centering
\setlength{\fboxsep}{0pt}
\begin{tabular}{c@{\hspace*{10pt}}c}
  \includegraphics[height=0.55in]{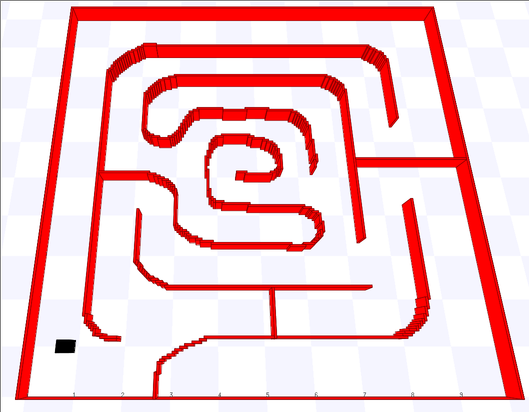} 
  \includegraphics[height=0.55in]{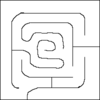} 
  \includegraphics[height=0.45in, width=0.45in]{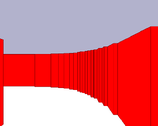} &
  \fbox{\includegraphics[height=0.65in]{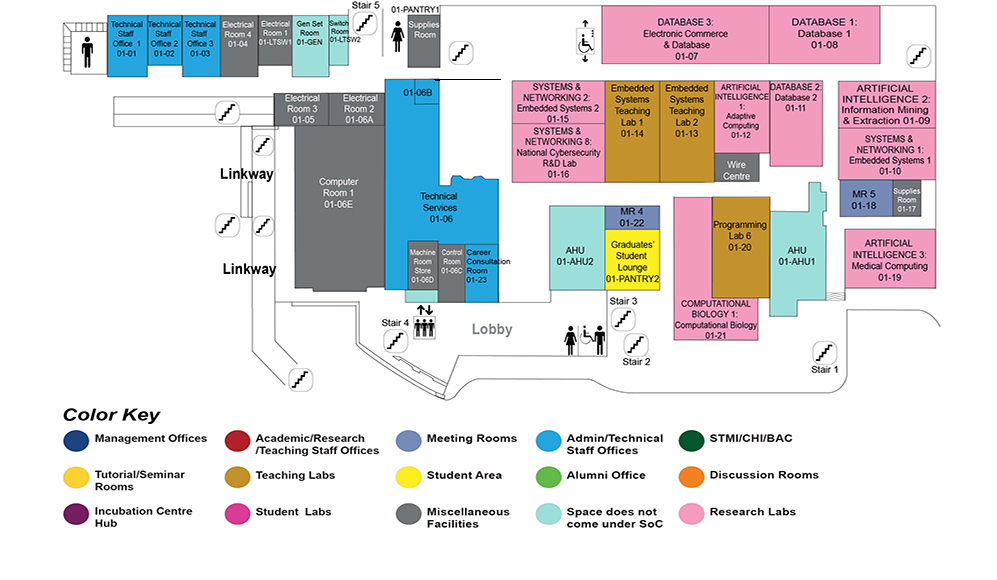}}
  \includegraphics[height=0.65in]{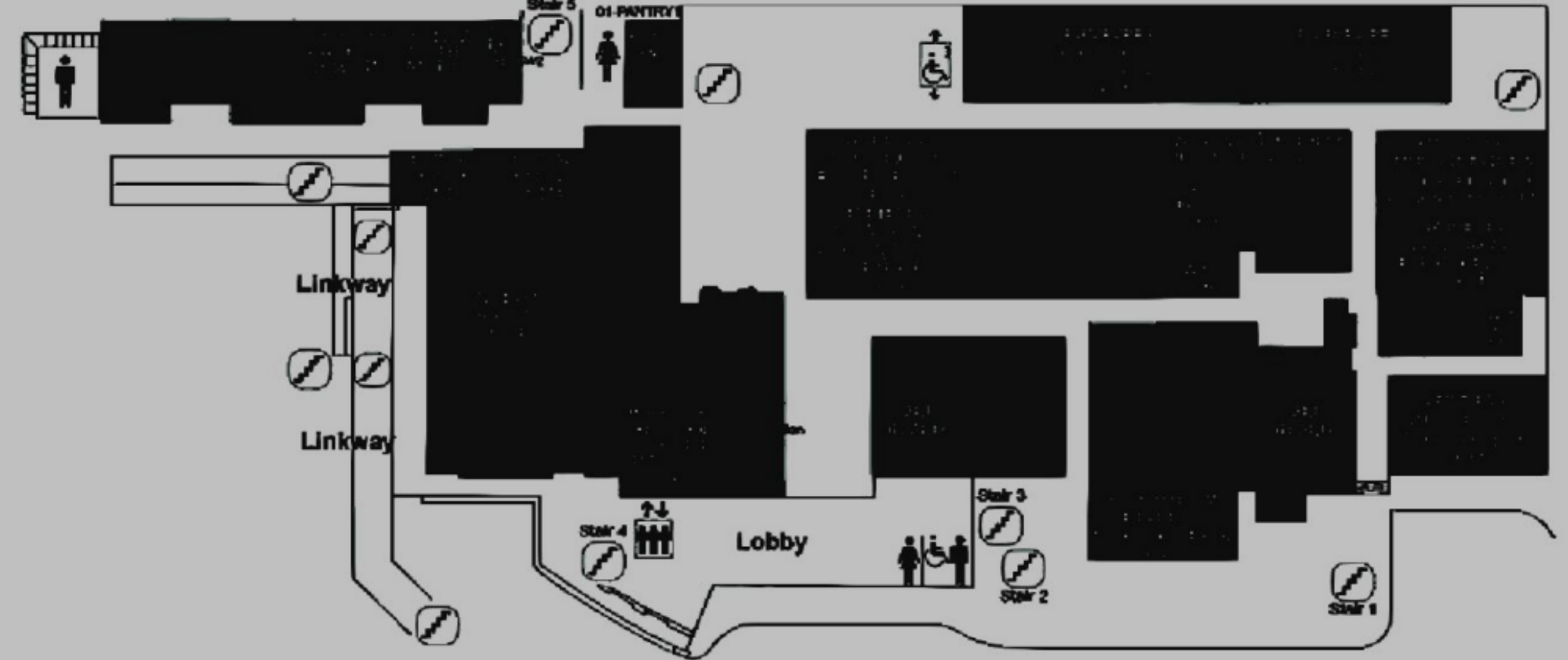}
  \includegraphics[height=0.5in]{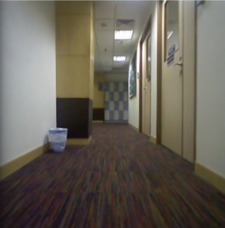}  \\
   {\small (\subfig a)} & {\small (\subfig b)} 
\end{tabular}
\caption{
Experimental setup in simulation and real-robot experiments. (\subfig a) A
simulated environment, the associated 2D map, and a camera view. (\subfig b) The
floor plan for a robot experiment environment, the
occupancy grid map computed from the floor plan, and a camera view of the environment.}
\label{fig:stage}
\end{figure}

We evaluated our approach in both simulation and real-robot experiments. For
simulation, we used the Stage simulator~\cite{vaughan2000stage} as the
experimental testbed. We hand-sketched a set of 2-D maps and used Stage to
generate the corresponding 3-D environments for training and testing
the intention-net. See \figref{fig:stage}\subfig a for an example. For the
real-robot experiment, we trained the robot on the first floor of our office
building and tested it on both the first floor and the second floor, which
differs significantly from the first floor geometrically and visually
(\figref{fig:real_performance}).  We digitized the visitor floor plans of our
building to  create crude occupancy-grid maps (\figref{fig:stage}\subfig
b).  Both the simulation and real-robot experiments used a Pioneer 3-DX
robot. In real-robot experiments, the robot is equipped with a webcam that has
$70^\circ$ field of view. It is also equipped with a Sick LMS-200 LIDAR, used for localization only.

For comparison, we consider three alternative methods: Path Tracker,
Non-Intention Net, and Dynamic Window~\cite{FoxBur97}. The first two are
ablated versions of our approach. Path Tracker replaces the intention-net with
a standard motion controller and follows the planned path without visual
feedback from the camera.  In contrast,
Non-Intention Net removes the intention as an input
to the neural-network controller and relies on visual feedback only. It
removes the path planner as well.  Dynamic Window is a well-established
successful method for robot navigation.
\subsection{Results from Simulation Experiments}

In simulation experiments, our main objective is to 
compare the two intention-net methods, DLM-Net and LPE-Net,
with the three alternative methods(\tabref{tab:sim_result}), 
according to four measures: success in task
completion, the number of human interventions, total task completion time, and
robot motion smoothness. \emph{Smoothness} is defined as the average
\emph{jerk}, \ie, the change in acceleration, of the path that the robot
traverses.

\begin{figure}[p]
\captionsetup[subfigure]{font=scriptsize,labelfont=scriptsize,position=top}
\vspace{-10px}
	\lineskip=0pt
	\centering
    \raisebox{-0.8\normalbaselineskip}[0pt][0pt]{\rotatebox[origin=c]{0}{\footnotesize A.\;}} 
    \subfloat[Task]{\includegraphics[width=0.14\textwidth]{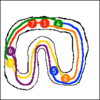}}%
    \hspace{2px}
    \subfloat[\IN]{\includegraphics[width=0.14\textwidth]{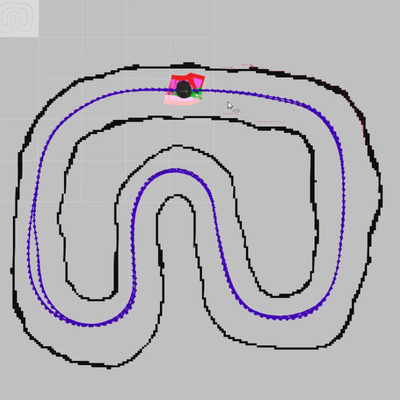}}%
    \hspace{2px}
    \subfloat[\IMN]{\includegraphics[width=0.14\textwidth]{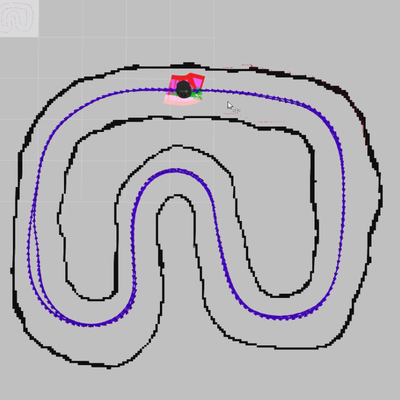}}%
    \hspace{2px}
    \subfloat[\BLIND]{\includegraphics[width=0.14\textwidth]{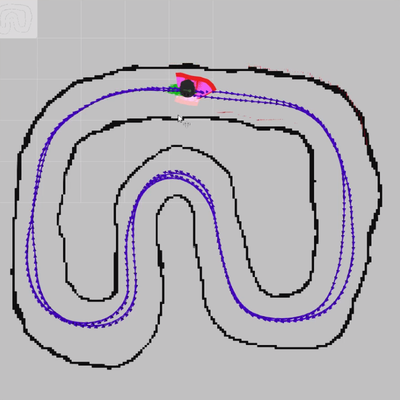}}%
    \hspace{2px}
    \subfloat[\PURE]{\includegraphics[width=0.14\textwidth]{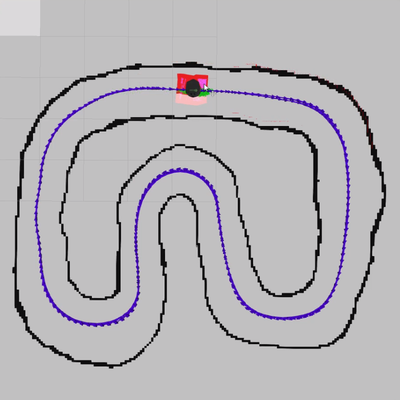}}%
    \hspace{2px}
    \subfloat[\DWA]{\includegraphics[width=0.14\textwidth]{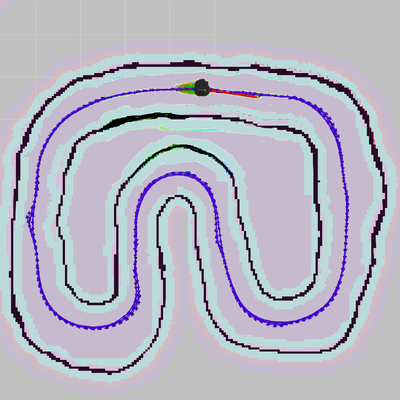}}%

    \vspace{3px}
    \raisebox{4.3\normalbaselineskip}[0pt][0pt]{\rotatebox[origin=c]{0}{\footnotesize B.\;}} 
    \includegraphics[width=0.14\textwidth]{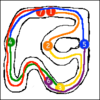}%
    \hspace{2px}
	\includegraphics[width=0.14\textwidth]{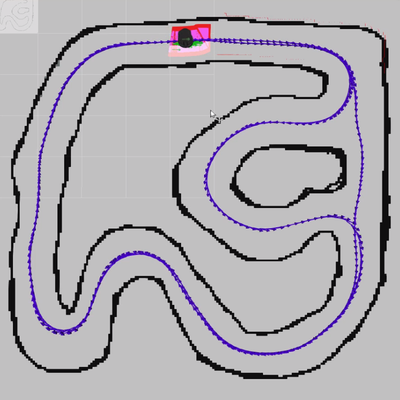}%
    \hspace{2px}
	\includegraphics[width=0.14\textwidth]{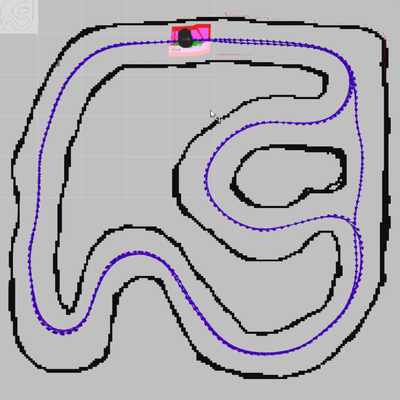}%
    \hspace{2px}
	\includegraphics[width=0.14\textwidth]{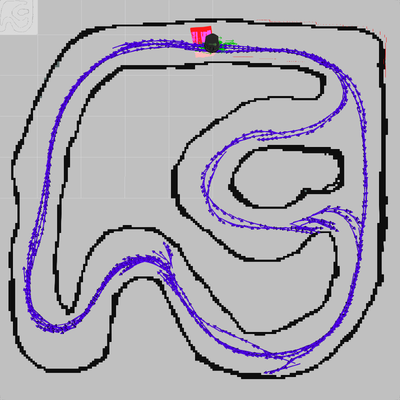}%
    \hspace{2px}
	\includegraphics[width=0.14\textwidth]{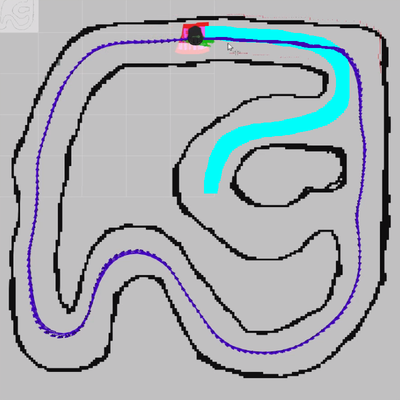}%
    \hspace{2px}
	\includegraphics[width=0.14\textwidth]{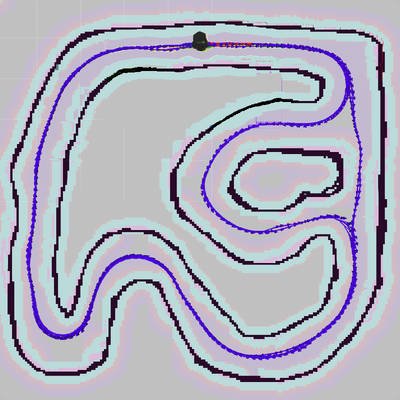}%

    \vspace{3px}
    \raisebox{4.3\normalbaselineskip}[0pt][0pt]{\rotatebox[origin=c]{0}{\footnotesize
        C.\;}}
	\includegraphics[width=0.14\textwidth]{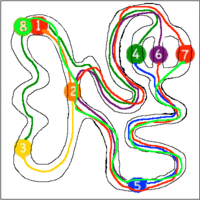}%
    \hspace{2px}
	\includegraphics[width=0.14\textwidth]{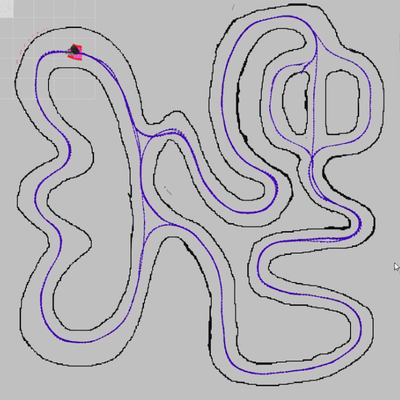}%
    \hspace{2px}
	\includegraphics[width=0.14\textwidth]{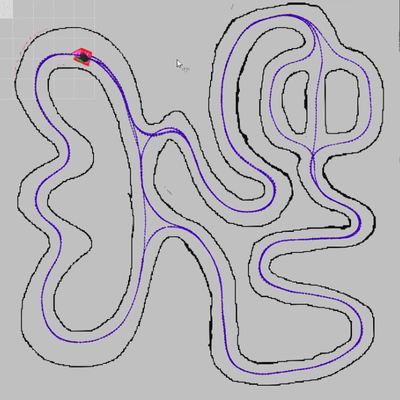}%
    \hspace{2px}
	\includegraphics[width=0.14\textwidth]{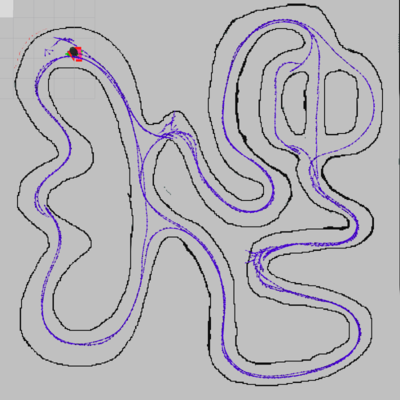}%
    \hspace{2px}
	\includegraphics[width=0.14\textwidth]{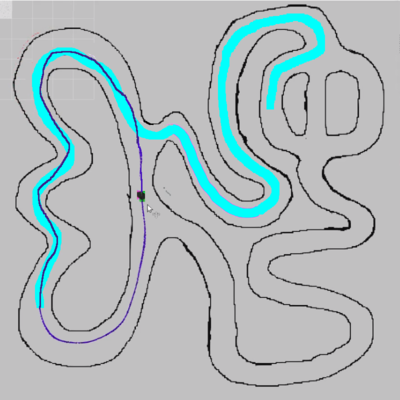}%
    \hspace{2px}
	\includegraphics[width=0.14\textwidth]{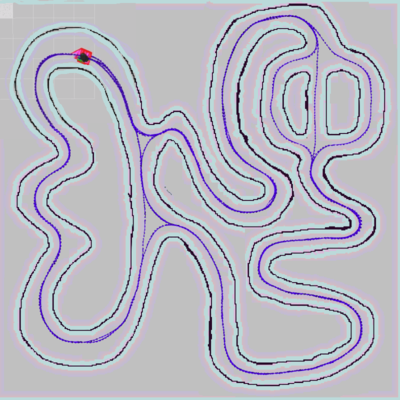}%

    \vspace{3px}
    \raisebox{4.3\normalbaselineskip}[0pt][0pt]{\rotatebox[origin=c]{0}{\footnotesize D.\;}} 
	\includegraphics[width=0.14\textwidth]{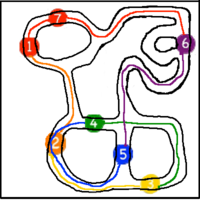}%
    \hspace{2px}
	\includegraphics[width=0.14\textwidth]{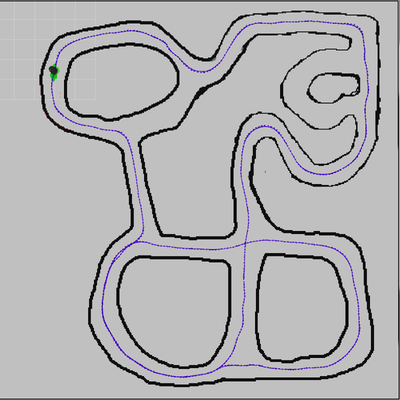}%
    \hspace{2px}
	\includegraphics[width=0.14\textwidth]{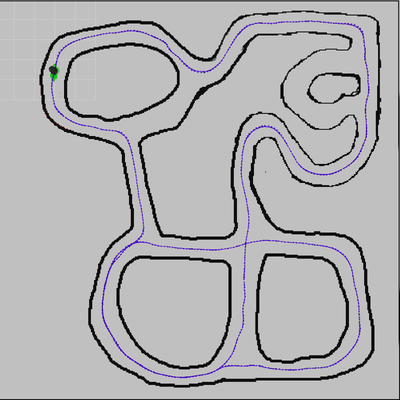}%
    \hspace{2px}
	\includegraphics[width=0.14\textwidth]{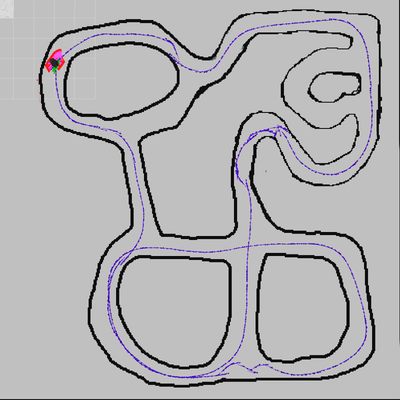}%
    \hspace{2px}
	\includegraphics[width=0.14\textwidth]{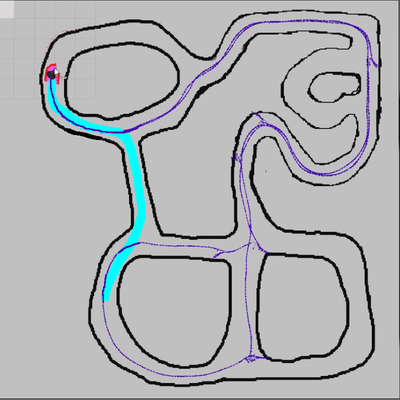}%
    \hspace{2px}
	\includegraphics[width=0.14\textwidth]{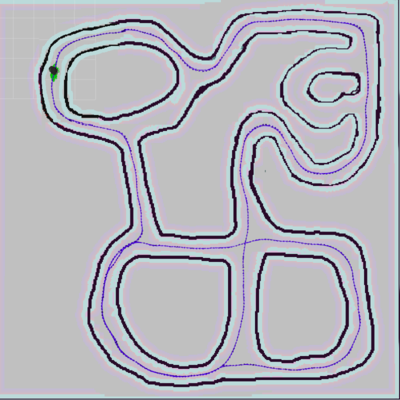}%

    \vspace{3px}
    \raisebox{4.3\normalbaselineskip}[0pt][0pt]{\rotatebox[origin=c]{0}{\footnotesize E.\;}} 
	\includegraphics[width=0.14\textwidth]{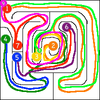}%
    \hspace{2px}
	\includegraphics[width=0.14\textwidth]{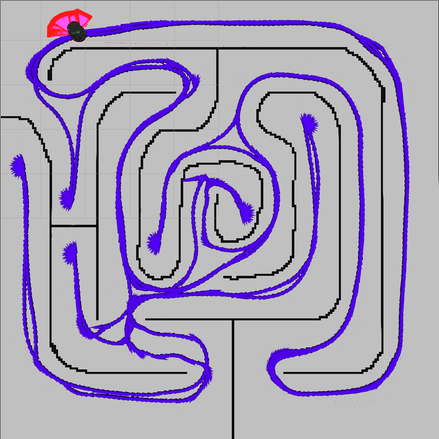}%
    \hspace{2px}
	\includegraphics[width=0.14\textwidth]{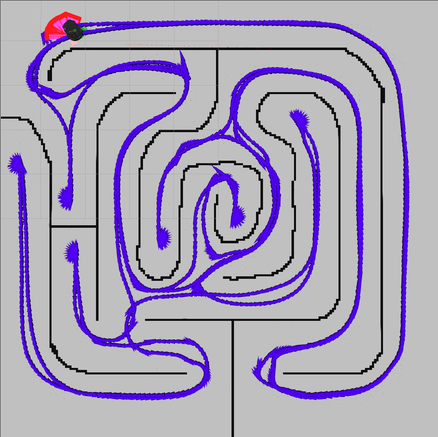}%
    \hspace{2px}
	\includegraphics[width=0.14\textwidth]{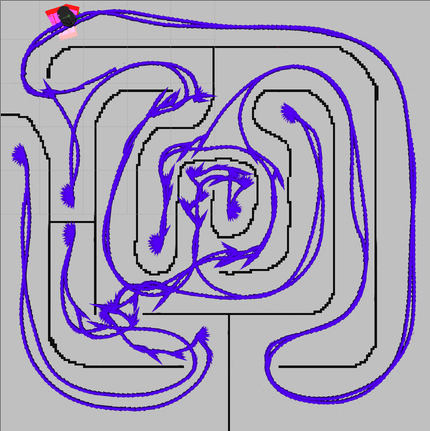}%
    \hspace{2px}
	\includegraphics[width=0.14\textwidth]{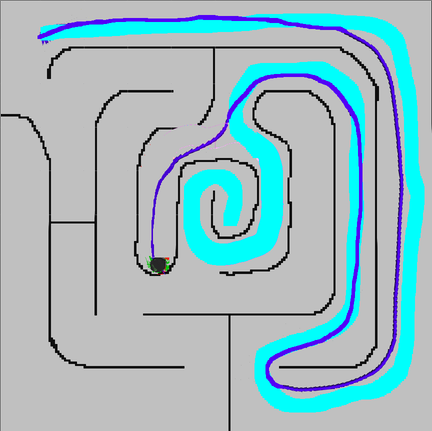}%
    \hspace{2px}
    \includegraphics[width=0.14\textwidth]{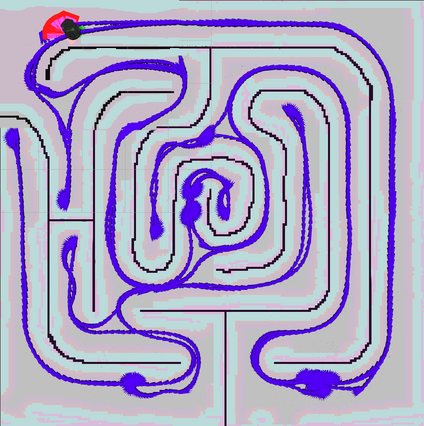}%
    \vspace*{6pt}
	\caption{Example runs of five navigation tasks in simulation. Each task
          requires the robot to navigate through a sequence of goals $1,2,
          \ldots$\,. The first column shows the task and the planned path. The
          remaining columns show the execution traces (thin blue lines)
          of five methods.
          For Non-Intention Net, we overlay the planned paths (thick
          light-blue lines) for the failed tasks. For tasks B, D, and E, it
          fails when traveling from goal 1 to 2. For task C, it fails when
          traveling from goal 3 to 4. 
        }
\label{fig:sim_qualitative}
\vspace{-20px}
\end{figure}

We consider five tasks (\tabref{tab:sim_result} and
\figref{fig:sim_qualitative}). Tasks A--C use environments present in the
training dataset and test for generalization to new goals. Tasks D--E test
for generalization to new environments unseen before.  The geometric complexity
of the environment increases from task A to E, roughly.  Task~A has a
simple environment with multi-way junctions. The environment  for task~E
is a maze.

\begin{table}[p]
  \caption{Performance comparison on five navigation tasks in simulation. Each
    task requires the robot to navigate through a sequence of goals. See \figref{fig:sim_qualitative}.}
\label{tab:sim_result}
\centering
\vspace*{6pt}
\scalebox{0.70}{
\begin{tabular}{clcccc}
  \hline

  \hline 

  Task & \multicolumn{1}{c}{Method} & Success & Intervention & Time (sec.) & Smoothness \\
  \hline
	 & \IN & Y & 0 & 113 & 0.0053 \\
	 & \IMN & Y & 0 & 114 & 0.0053 \\
A & \BLIND & Y & 0 & \textbf{105} & 0.0040\\
	 & \PURE & Y & 0 & 112 & 0.0053\\
	 & \DWA & Y & 0 & 109 & \textbf{0.0039} \\
\hline
	& \IN & Y & 0 & \textbf{118} & 0.0074\\
	& \IMN & Y & 0 & 128 & \textbf{0.0068} \\
B & \BLIND & Y & 3 & 155 & 0.0170 \\
    & \PURE & N & -- & -- & -- \\
	& \DWA & Y & 0 & 126 & 0.0100 \\
\hline
	& \IN & Y & 0 & \textbf{559} & 0.0074 \\
	& \IMN & Y & 0 & 561 & \textbf{0.0072} \\
C & \BLIND & Y & 16 & 640 & 0.0150 \\
    & \PURE & N & -- & -- & -- \\
    & \DWA & Y & 0 & 565 & 0.0094 \\
\hline
	& \IN & Y & 0 & 237 & 0.0085 \\
	& \IMN & Y & 0 & \textbf{226} & \textbf{0.0066} \\
D & \BLIND & Y & 5 & 240 & 0.0120 \\
    & \PURE & N & -- & -- & -- \\
	 & \DWA & Y & 0 & 238 & 0.0095 \\
\hline
	& \IN & Y & 0 & 545 & 0.0080 \\
	 & \IMN & Y & 0 & \textbf{531} & \textbf{0.0075} \\
E & \BLIND & Y & 21 & 546 & 0.0089 \\
    & \PURE & N & -- & -- & -- \\
	 & \DWA & Y & 0 & 551 & 0.0084 \\
\hline
\end{tabular}}
\vspace{-30px}
\end{table}


Overall LPE-Net performs better than DLM-Net, especially in task D and
E, which involve generalization to new environments. Both substantially
outperform the alternatives. They complete all tasks with no human
interventions. Path Tracker completes the task faster than others in the
simple task, task A. However, it generally produces robot motions that are
jerky and not smooth.  This is quite visible in \figref{fig:sim_qualitative}.
It also requires a large number of human interventions in the most difficult
task, task E.  Non-Intention Net completes task A, but fails in all
others. This is unsurprising. By removing the intention input, Non-Intention
Net has no notion of global navigation goals.  It blindly mimics the expert,
without recognizing that given the same visual feedback, the expert may have
different goals in the training dataset. It succeeds in task A, because this
task has a very simple environment, with no junction to separate paths for
different goals. LPE-Net and DLM-Net also outperform Dynamic Window in terms
of task completion time and robot motion smoothness, especially in the more
complex environments (tasks C--E).

\subsection{Results from Real-Robot  Experiments}
\label{result:realworld}



Next, we evaluate the performance of our
approach, when faced with the complexity of the real world.  We highlight
several interesting scenarios below:
\begin{figure}
\captionsetup[subfigure]{font=scriptsize,labelfont=scriptsize,position=top}
	\centering
    \raisebox{-2.8\normalbaselineskip}[0pt][0pt]{\rotatebox[origin=c]{0}{\small
        (a)\hspace*{3pt}}} 
    \subfloat[DLM-Net]{
	\includegraphics[width=0.15\textwidth]{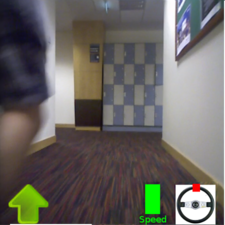}%
    \hspace{1px}
    \includegraphics[width=0.15\textwidth]{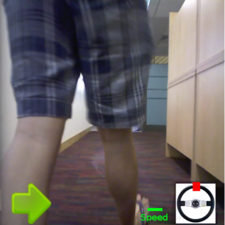}}%
    \hspace{2.5px}
    \subfloat[LPE-Net]{
	\includegraphics[width=0.15\textwidth]{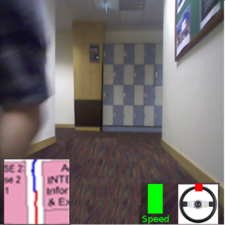}%
    \hspace{1px}
    \includegraphics[width=0.15\textwidth]{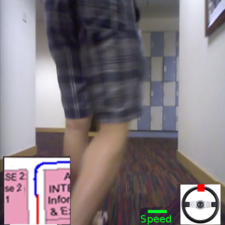}}%

    \vspace{2px}
    \raisebox{2.8\normalbaselineskip}[0pt][0pt]{\rotatebox[origin=c]{0}{\small
        (b) \hspace*{3pt}}} 
	\includegraphics[width=0.15\textwidth]{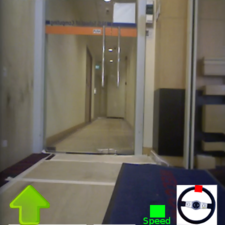}%
    \hspace{1px}
	\includegraphics[width=0.15\textwidth]{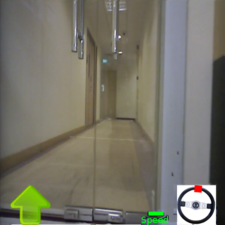}%
    \hspace{5px}
	\includegraphics[width=0.15\textwidth]{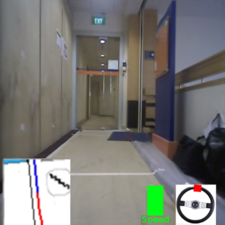}%
    \hspace{1px}
	\includegraphics[width=0.15\textwidth]{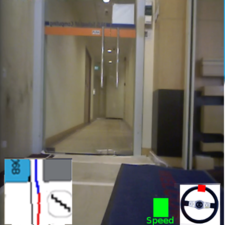}%

    \vspace{2px}
    \raisebox{2.8\normalbaselineskip}[0pt][0pt]{\rotatebox[origin=c]{0}{\small
        (c) \hspace*{3pt}}} 
	\includegraphics[width=0.15\textwidth]{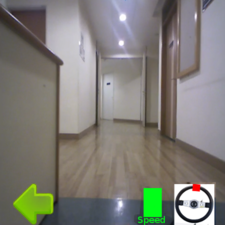}%
    \hspace{1px}
	\includegraphics[width=0.15\textwidth]{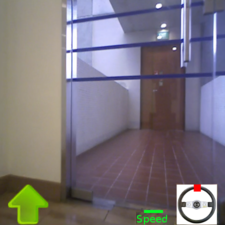}%
    \hspace{5px}
	\includegraphics[width=0.15\textwidth]{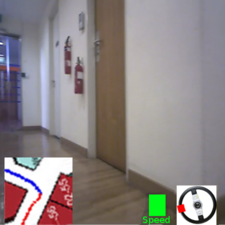}%
    \hspace{1px}
	\includegraphics[width=0.15\textwidth]{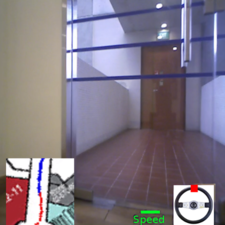}%
	\caption{Examples from real-robot experiments. (\subfig a) The robots
          slow down to let a human blocking its path pass by.
          (\subfig b) The robot encounters a glass door in the training dataset.  (\subfig c) The robot encounters a different glass door not
        in the training dataset. The robot detects the glass door and slows
        down to a stop, as indicated by the speedometer at the lower right
        corner of each image.}
\label{fig:real_qualitative}
\end{figure}

\begin{figure}
\centering
\begin{tabular}{c@{\hspace*{1pt}}c@{\hspace*{1pt}}cc@{\hspace*{1pt}}c@{\hspace*{1pt}}c}
    \includegraphics[height=0.75in]{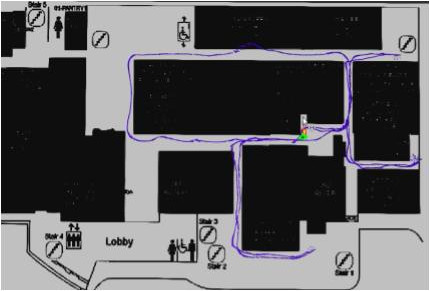}&
    \includegraphics[height=0.75in,width=0.75in]{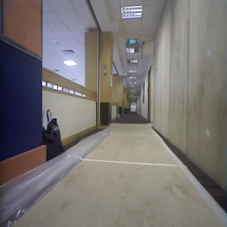}&
    \includegraphics[height=0.75in,width=0.75in]{figures/firstfloor_visual1.png}&
    \includegraphics[height=0.75in]{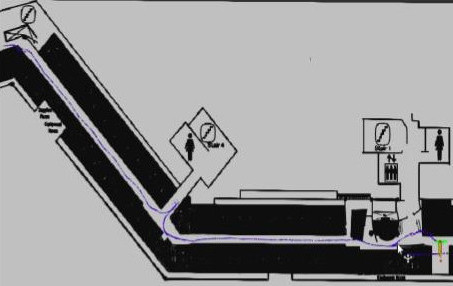}
    \includegraphics[height=0.75in,width=0.75in]{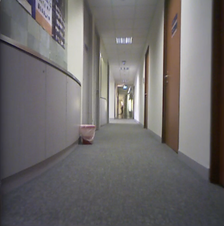}&
    \includegraphics[height=0.75in,width=0.75in]{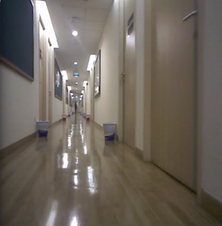}&
  \\
  \multicolumn{3}{c}{\small (\subfig a)} & \multicolumn{3}{c}{\small (\subfig b)} 
\end{tabular}
    \caption{Generalization to a new environment. (\subfig a) The first floor
      of the building in robot experiments. The robot is trained on data from
      the first
      floor only. (\subfig b) The second floor, used for testing the robot, has
      a significantly different geometric arrangement and visual appearance. 
} \label{fig:real_performance}
\end{figure}

\begin{itemize}
\item \textit{Moving people.}  The crude floor-plan maps obviously do not
  contain information on dynamic obstacles, such as moving people. Nevertheless,
  our approach can handle people obstructing the robot path, thanks to the
  intention-net learned from data.  The robot recognizes people, men or women
  dressed in different types of clothing.  If a person moves in the same
  direction ahead of the robot, the robot usually slows down and follows the
  people. If a person moves in the opposite direction, the robot stops and
  waits for the people to pass by.  See \figref{fig:real_qualitative}\subfig a
  for an example.

\item \textit{Glass doors.} Transparent obstacles such as glass doors present
  a well-known difficulty for navigation systems that rely on LIDAR.  Using
  images from a single monocular webcam, the intention-net enables our
  robot to handle glass doors. \figref{fig:real_qualitative}\subfig b shows
  the glass door in the training dataset. \figref{fig:real_qualitative}\subfig c shows the robot behavior when it
  encounters a different glass door not present in the dataset.  Even more
  interestingly, the floor-plan map is inaccurate and 
  does not contain the glass door. So
  the high-level path planner is not aware of it. The intention-net
  nevertheless detects the glass door and responds to it correctly by slowing
  down the robot to a full stop. 

\item \textit{New environments.} The robot is trained on data from the first
  door of our office building only. It navigates very capable on the second
  floor, which differs significantly from the first floor in geometric
  arrangement and visual appearance (\figref{fig:real_performance}), though
  more thorough evaluation is required to quantify the performance. 
\end{itemize}


Both \IN and \IMN are evaluated in our experiments. Overall they both perform
well, but \IMN usually outperforms \IN when difficult maneuvers,
such as sharp turns, are required.  \IN restricts the intention to four
discrete values based on a manually chosen threshold.
While it is possible to learn the threshold from data, we believe that \IMN,
which uses a richer intention representation, is a better and more systematic
way to proceed. 

The most common failure of our approach occurs when the planned path generates
the wrong intention as a result of inaccurate localization, especially with
DLM-Net.  For example, the robot approaches the end of a hallway and must turn
left.  Intuitively, the \cmd{TurnLeft} intention is required. Instead, the
\cmd{GoForward} is generated because of the localization error. The robot may
then crash against the wall. This happens because the simple $A^*$ path planner
does not account for localization errors and relies solely  on replanning
to close the loop. A more sophisticated planner, which hedges against
uncertainty in localization, would alleviate this difficulty.


\begin{figure}[t]
	\lineskip=0pt
    \centering
	\includegraphics[height=0.08\textwidth,width=0.08\textwidth]{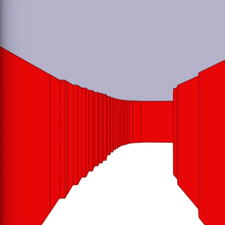}\hspace{0.5pt}%
	\includegraphics[height=0.08\textwidth,width=0.08\textwidth]{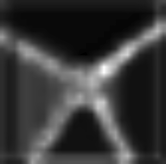}\hspace{0.5pt}%
	\includegraphics[height=0.08\textwidth,width=0.08\textwidth]{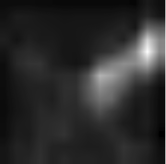}\hspace{0.5pt}%
	\includegraphics[height=0.08\textwidth,width=0.08\textwidth]{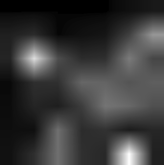}%
	\hspace{1.5pt}
	\includegraphics[height=0.08\textwidth,width=0.08\textwidth]{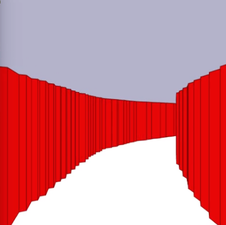}\hspace{0.5pt}%
	\includegraphics[height=0.08\textwidth,width=0.08\textwidth]{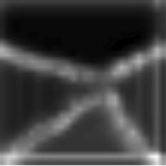}\hspace{0.5pt}%
	\includegraphics[height=0.08\textwidth,width=0.08\textwidth]{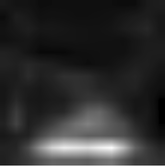}\hspace{0.5pt}%
	\includegraphics[height=0.08\textwidth,width=0.08\textwidth]{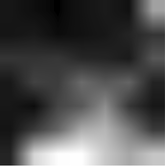}%
	\hspace{1.5pt}
	\includegraphics[height=0.08\textwidth,width=0.08\textwidth]{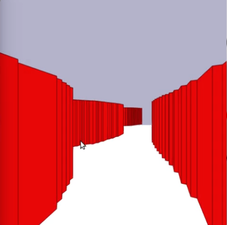}\hspace{0.5pt}%
	\includegraphics[height=0.08\textwidth,width=0.08\textwidth]{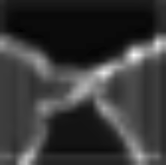}\hspace{0.5pt}%
	\includegraphics[height=0.08\textwidth,width=0.08\textwidth]{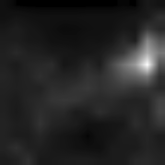}\hspace{0.5pt}%
	\includegraphics[height=0.08\textwidth,width=0.08\textwidth]{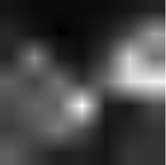}%

	\vspace{5px}
	\includegraphics[height=0.08\textwidth,width=0.08\textwidth]{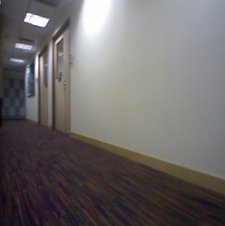}\hspace{0.5pt}%
	\includegraphics[height=0.08\textwidth,width=0.08\textwidth]{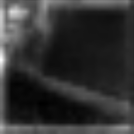}\hspace{0.5pt}%
	\includegraphics[height=0.08\textwidth,width=0.08\textwidth]{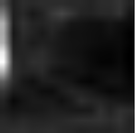}\hspace{0.5pt}%
	\includegraphics[height=0.08\textwidth,width=0.08\textwidth]{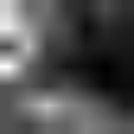}%
	\hspace{1.5pt}
	\includegraphics[height=0.08\textwidth,width=0.08\textwidth]{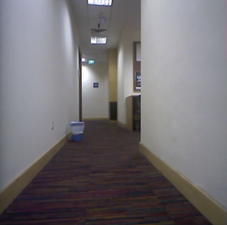}\hspace{0.5pt}%
	\includegraphics[height=0.08\textwidth,width=0.08\textwidth]{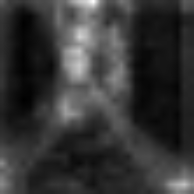}\hspace{0.5pt}%
	\includegraphics[height=0.08\textwidth,width=0.08\textwidth]{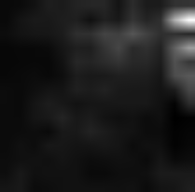}\hspace{0.5pt}%
	\includegraphics[height=0.08\textwidth,width=0.08\textwidth]{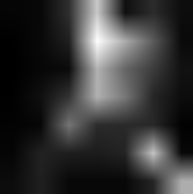}%
	\hspace{1.5pt}
	\includegraphics[height=0.08\textwidth,width=0.08\textwidth]{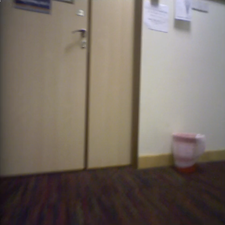}\hspace{0.5pt}%
	\includegraphics[height=0.08\textwidth,width=0.08\textwidth]{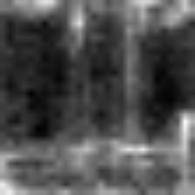}\hspace{0.5pt}%
	\includegraphics[height=0.08\textwidth,width=0.08\textwidth]{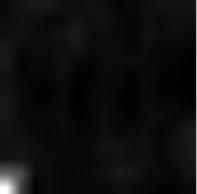}\hspace{0.5pt}%
	\includegraphics[height=0.08\textwidth,width=0.08\textwidth]{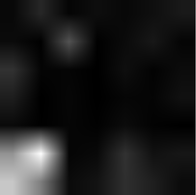}%
	\caption{Visualization of the  learned feature in simulation 
          and real-robot experiments. Each block shows the camera
          view and the activated features from the 9th, 18th, and 50th layers
          of ResNet50.
        }
\label{fig:visualization}
\end{figure}

\subsection{Visualization of Learned Features}

To gain some understanding of how the intention-net processes the visual
feedback, we visualize the learned features by projecting all activated
features of a layer on an image.
\figref{fig:visualization} shows some examples.
The lower-layer features are more concrete and capture the lines where the wall
meets the floor or the ceiling. The is clear, especially in the simulation
setting. The  higher-layer features are more abstract and difficult to
interpret. Sometimes they capture the free space for navigation, \eg, the
lower-left block of \figref{fig:visualization}.

\section{Conclusion}
\label{sec:conclusion}
We propose a two-level hierarchical approach that integrates
model-free deep learning and model-based path planning for reliable robot
navigation. Learning allows us to train the intention-net, a neural-network
motion controller, to be robust against robot perceptual noise and
localization errors. Planning enables the learned controller to generalize to
new environments and goals effectively. Together they enable our robot to
navigate almost immediately in a new indoor environment, with a very crude
map, such as a digitized visitor floor plan.

One main limitation of our current approach is robot localization. It treats
localization as a black box and segregates it from the planner. Integrating
localization with planning and learning will be the next challenge in our
work.




\clearpage 
\acknowledgments{This research is supported in part by Singapore Ministry of
  Education AcRF grant MOE2016-T2-2-068 and Singapore-MIT Alliance for
  Research \& Technology IRG grant R-252-000-655-592. We thank Bin Zhou from
  the Panasonic R\&D Center Singapore for helping with some experiments.
}


\bibliography{hippocampus17a}  

\end{document}